\documentclass[letterpaper, 10pt, conference]{ieeeconf}
 
\IEEEoverridecommandlockouts
\usepackage{cite}
 
\usepackage{graphicx}
\graphicspath{{figures/}}
\DeclareGraphicsExtensions{.pdf,.png,.jpg,.jpeg}
 
\usepackage{amsmath}
\usepackage{amssymb}
 
\usepackage{algorithmic}
\usepackage{array}
\usepackage{booktabs}
\usepackage{multirow}
 
\usepackage[font=footnotesize]{subfig}

\usepackage[font=footnotesize,labelfont=bf,labelsep=colon,justification=justified,singlelinecheck=false,skip=4pt]{caption}
\PassOptionsToPackage{hyphens}{url}
\usepackage{url}
\usepackage[hidelinks]{hyperref}

\hypersetup{
  pdftitle={OPTED: On-Policy Fine-Tuning for End-to-End Driving using a Render-Free Teacher},
  pdfauthor={Damiano Da Col, Maximilian Igl, Peter Karkus, Kashyap Chitta, Boris Ivanovic, Marco Pavone, Konrad Schindler, Christos Sakaridis},
  pdfsubject={Autonomous driving, on-policy distillation, closed-loop fine-tuning},
  pdfkeywords={autonomous driving, end-to-end driving, on-policy distillation, reinforcement learning, imitation learning, closed-loop fine-tuning}
}
 
\def\tablescaler{0.9}

\title{\LARGE \bf
OPTED: On-Policy Fine-Tuning for End-to-End Driving\\using a Render-Free Teacher
}
 
\author{Damiano Da Col$^{1,2}$, Maximilian Igl$^{3}$, Peter Karkus$^{3}$, Kashyap Chitta$^{1,4*}$, \\ 
Boris Ivanovic$^{3}$, Marco Pavone$^{3,5}$, Konrad Schindler$^{2}$, and Christos Sakaridis$^{2}$%
\thanks{$^{1}$KE:SAI. $^{2}$ETH Z\"urich. $^{3}$NVIDIA Research. $^{4}$ELLIS Institute T\"ubingen. $^{5}$Stanford University. $^{*}$Work done in part while at NVIDIA Research. Contact: {\tt\small damiano@kesai.eu}}%
}
 
\begin{document}
 
\maketitle
\thispagestyle{empty}
\bstctlcite{IEEEexample:BSTcontrol}
\pagestyle{empty}

\begin{abstract}
As scaling pre-training data alone yields diminishing returns, post-training is becoming increasingly important across physical AI domains such as autonomous driving. End-to-end driving policies are pre-trained in open loop with behavior cloning on human demonstrations. However, compounding errors during closed-loop deployment can take the vehicle outside the training data distribution, increasing the risk of safety-critical incidents. Closed-loop post-training can mitigate this risk but requires costly simulation for sensor-based policies. We propose OPTED (on-policy fine-tuning for end-to-end driving) which decouples reinforcement learning from the post-training of the end-to-end policy: a privileged teacher is trained using RL on vectorized inputs (HD-map and bounding boxes). This teacher then provides supervision to the pre-trained student during closed-loop post-training. We apply OPTED to two camera-based models, TransFuser and VaVAM, and fine-tune them in AlpaSim, using neural reconstructions (3DGS) of real driving logs. Driving scores increase by factors of 1.6$\times$ and 9.5$\times$, respectively. In controlled experiments OPTED matches closed-loop performance with approximately three orders of magnitude fewer simulator interactions than direct RL post-training, while staying closer to the human prior.

\noindent Project page: \url{https://01dami23.github.io/opted/}
\end{abstract}

\section{Introduction}
\label{sec:intro}

End-to-end driving policies are pre-trained with behavior cloning (BC) to map raw sensor data to actions or waypoint trajectories~\cite{Wu2022NIPS,Nguyen2026CVPR,Bartoccioni2025ARXIV}. As in other domains of artificial intelligence, where scaling pre-training data alone yields diminishing returns, post-training is becoming critical for raising a driving policy's capability~\cite{Karkus2025TECHRXIV,Li2026ARXIVa}. In particular, because BC is an open-loop objective, the policy is only supervised on states visited by the demonstrator.
In contrast, during closed-loop deployment of the policy, its own actions determine the future states it observes. Consequently, small errors can compound over time, and the policy may reach states that are not well covered by the dataset~\cite{Ross2010AISTATS}. Closed-loop post-training addresses this gap.

One such approach, reinforcement learning (RL), can directly optimize safety and progress in closed loop. For example, using efficient simulation on vectorized map and bounding box inputs, RL trained on billions of interactions has achieved infraction rates below those of human drivers~\cite{Cusumano-Towner2025ICML}. 
In contrast, sensor-based policies require more expensive 3D rendering for high-fidelity simulations, and post-training on large amounts of rendered frames increases the risk of overfitting to simulation.
Consequently, despite growing interest, practical RL post-training of end-to-end models at the scale required for superhuman driving is yet to be demonstrated~\cite{Gao2025NIPS,Wang2025ARXIV}.
Among more sample-efficient alternatives to RL, closed-loop supervised fine-tuning uses human logs for supervision; however, the supervision becomes increasingly biased as the policy deviates from logged trajectories~\cite{Zhang2025CVPR,Garcia-Cobo2026CVPRFIND}. DAgger~\cite{Ross2011AISTATS} instead labels states visited by the trained policy with expert actions, but requires an expert policy that can be queried at scale.

\begin{figure}[t]
\centering
\includegraphics[width=\columnwidth]{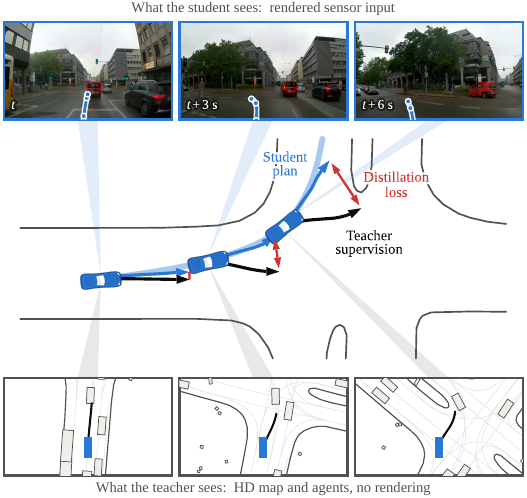}
\caption{\textbf{OPTED.} A pre-trained end-to-end student is rolled out in closed loop in a sensor-based simulator (\emph{top:} the rendered views the student consumes). At visited states, a privileged teacher, trained with reinforcement learning (RL) on vectorized HD-map observations in a render-free simulator (\emph{bottom}), provides the student's distillation target. Rendered interaction is limited to the end-to-end model, while RL exploration, which is computationally infeasible with rendering, is confined to the render-free simulator.}
\label{fig:teaser}
\vspace{-0.5cm}
\end{figure}

\begin{figure*}[t]
\centering
\includegraphics[width=\textwidth]{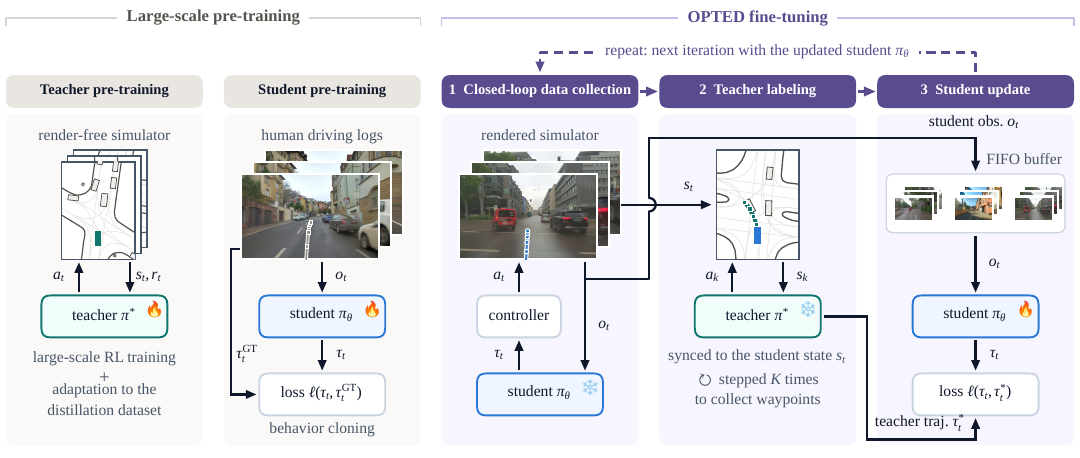}
\caption{\textbf{The OPTED pipeline.} \emph{Large-scale pre-training} produces the privileged teacher trained with RL in a render-free simulator on vectorized states (HD-map polylines and actor bounding boxes), subsequently adapted with further RL training to the distillation dataset. The end-to-end student is trained in open loop via behavior cloning on images recorded from human demonstrations, regressing its plan $\tau_t$ onto the logged trajectory $\tau_t^{\mathrm{GT}}$. OPTED fine-tuning then runs three steps iteratively. (1)~The student drives the rendered simulator in closed loop: it receives frame $o_t$, predicts $\tau_t$, and the controller converts the plan into the action $a_t$ that advances the simulator. (2)~Each visited state is synchronized into the vectorized simulator, where the frozen teacher is stepped $K$ times to produce a trajectory of length $|\tau_t|$. (3)~Student observations fill a bounded FIFO buffer of recent rollouts, and the student is trained on them with the imitation loss $\ell$, KL for action heads or weighted $\mathrm{L}_1$ for waypoint heads (Eqs.~\ref{eq:kl-loss},~\ref{eq:traj-loss}). }
\label{fig:pipeline}
\vspace{-0.5cm}
\end{figure*}

We hypothesize that the two costs of closed-loop post-training, sample-inefficient exploration and sample-efficient supervision, can be decoupled and incurred in different simulators. Exploration does not require sensor input, so an RL policy trained without rendering can serve as the expert for on-policy supervision, provided the two simulators are synchronized to the same state.
Based on this, we propose \emph{On-Policy fine-Tuning for End-to-end Driving} (OPTED),
illustrated in Fig.~\ref{fig:teaser}.
A smaller, privileged teacher is trained with RL on vectorized inputs and then frozen. A larger, pre-trained student utilizing RGB images as input is subsequently rolled out in simulation supporting the required image rendering.
During those more expensive rollouts, each visited state is synchronized into the teacher's simulator and labeled with the teacher's behavior. Importantly, by decoupling the teacher and student into separate simulators, we allow both to be pre-trained on large, unpaired, independent collections of data. This makes our approach more scalable and more widely applicable than distillation within a single simulator~\cite{Chen2019CORL,Zhang2021ICCV}.

Furthermore, state-of-the-art RL policies output actions~\cite{Cusumano-Towner2025ICML,Kazemkhani2025ICLR,Jaeger2025CORL}, whereas end-to-end policies predominantly predict waypoint trajectories~\cite{Wu2022NIPS,Nguyen2026CVPR,Bartoccioni2025ARXIV,Wang2025ARXIV}. OPTED bridges this mismatch by rolling out the teacher from the student's state and converting the resulting motion into a target trajectory in the student's frame. No constraint is placed on the student's architecture or output representation; the distillation loss is adapted to the student's output representation.
OPTED thus combines the closed-loop competence of large-scale RL on vectorized inputs with the human-data prior of existing end-to-end models, with supervision applied on the student's own rollouts in reconstructions of real sensor logs.

\noindent \textbf{Contributions.} (i)~We propose OPTED, a simple and general method for closed-loop post-training of driving policies, in which a privileged teacher trained with render-free RL provides supervision to closed-loop rollouts of a student acting on visual inputs. Our setup bridges different output representations between teacher and student, in particular, imposing no restrictions on the student's model architecture or output representation. (ii)~We apply OPTED to two architecturally diverse public models, TransFuser (LTFv6)~\cite{Nguyen2026CVPR} and VaVAM~\cite{Bartoccioni2025ARXIV}, in the AlpaSim~\cite{alpasim_2025} simulator on neural reconstructions of real logs~\cite{NVIDIA2025HF}, improving their closed-loop scene scores by factors of 1.6$\times$ and 9.5$\times$ respectively, after two epochs of fine-tuning, with only a moderate increase in open-loop displacement error from the logged human trajectories. (iii)~We provide additional insights through a controlled study on vectorized policies in PufferDrive~\cite{Cornelisse2025GH} on WOMD~\cite{Ettinger2021ICCV}, a setting in which RL fine-tuning can be run to convergence. We show that OPTED matches the closed-loop performance of RL fine-tuning with approximately three orders of magnitude fewer closed-loop simulator interactions while staying closer to the human prior than RL fine-tuning.

\section{Related Work}

\noindent \textbf{Reinforcement Learning for Driving.}
\label{related_work:reinforcement_learning}
RL optimizes closed-loop objectives directly and requires no demonstrated targets, at the cost of extensive exploration. On privileged inputs, whether vectorized~\cite{Kazemkhani2025ICLR,Cornelisse2025GH} or rasterized BEV~\cite{Jaeger2025CORL}, simulators execute billions of transitions and self-play policies attain infraction rates below those of human drivers~\cite{Cusumano-Towner2025ICML}; while RL fine-tuning has been demonstrated to improve pre-trained vectorized planners~\cite{Pei2026ICLR,Li2026CVPR,Tang2026ICLR}. On sensor inputs, every interaction must be rendered. Existing approaches either train only on synthetic renderings of simulator primitives rather than real images~\cite{Yin2026ARXIV}, or render reconstructions of real scenes, but are constrained by the higher resource requirements of this approach~\cite{Gao2025NIPS,Gao2026ARXIV,Wang2025ARXIV}. OPTED confines RL to the vectorized state, where interaction is cheapest, and transfers the resulting policy to the sensor domain through supervision. In a controlled setting where both can be run to convergence, OPTED matches RL fine-tuning of the same student with three orders of magnitude fewer closed-loop episodes (Sec.~\ref{sec:exp:womd}).

\noindent \textbf{Closed-Loop Supervised Fine-Tuning.}
\label{related_work:supervised_fine_tuning}
A second line of work replaces the reward with supervision derived from recorded human driving, either by perturbing logged trajectories offline~\cite{Bansal2019RSS} or by rolling out the policy in closed loop and training it toward its own sampled prediction closest to the log~\cite{Zhang2025CVPR}. RoaD~\cite{Garcia-Cobo2026CVPRFIND} applies the latter procedure to a sensor-based policy in AlpaSim~\cite{alpasim_2025}. Because the target is selected by proximity to the logged trajectory and recovery brings the state back to it, the training distribution remains anchored to the demonstrated states rather than to those the policy reaches when left to drift. We compare against closed-loop SFT in Sec.~\ref{sec:exp:end-to-end}.

\noindent \textbf{Privileged Distillation for Driving.}
\label{related_work:privileged_distillation}
End-to-end policies map raw sensor data to actions or waypoints~\cite{Wu2022NIPS} and are predominantly trained with BC on human demonstrations. BC supervises only demonstrated states, while during closed-loop deployment errors compound and the policy can reach states from which it has not learned to recover~\cite{Ross2010AISTATS}. DAgger~\cite{Ross2011AISTATS} corrects this by querying an expert at the states the learner visits, but requires an oracle that can label arbitrary states. Privileged distillation provides such an oracle through a learned teacher that observes the privileged state, and supervises the sensor-based student either on the states it visits~\cite{Chen2019CORL,Zhang2021ICCV} or offline on a fixed corpus of teacher rollouts~\cite{Wu2022NIPS,Jia2023ICCV}. In these pipelines, teacher and student share one simulator, and the student is designed jointly with the pipeline and trained from scratch. LEAD~\cite{Nguyen2026CVPR} instead restricts a rule-based expert's supervision to what the student can perceive. Concurrent work distills vectorized teachers into pixel-based students: Gigapixel~\cite{Rowe2026ARXIV} runs self-play DAgger in a stylized renderer and reaches real images only through an offline perception adapter with the planning head frozen, whereas TerraTransfer~\cite{Xiong2026ARXIV} distills offline on logged human states and never supervises the student on its own rollouts. OPTED differs in three respects. The teacher is trained in a separate render-free simulator on an independent vectorized corpus, so exploration is not bounded by the cost of the student's simulator, and it interfaces with the student only through the HD map. Its labels are applied to the student's own closed-loop rollouts rendered from reconstructions of real logs, so the student is supervised on the states it actually reaches with realistic sensor inputs. The student is a pre-trained model fine-tuned with its output head unchanged, so its human-data prior and architecture are retained.

\section{Method}
\label{sec:method}

The high computational cost of closed-loop RL fine-tuning of an end-to-end driving policy is due to two factors: (i) RL is sample-inefficient and requires a large number of exploration steps; (ii) each exploration step requires sensor simulation and therefore expensive rendering~\cite{Wang2025ARXIV,Gao2025NIPS}. In this section we introduce OPTED, which decouples the two~(Fig.~\ref{fig:pipeline}).

\begin{table*}[t]
\centering
\caption{Closed-loop evaluation on the 441 held-out NuRec scenes~\cite{NVIDIA2025HF} with AlpaSim challenge metrics and further sub-metrics, open-loop ADE on NuRec and nuPlan, and inference latency per planning step. Bold marks the best value per architecture.}
\label{tab:e2e_main_results}
\small
\setlength{\tabcolsep}{3pt}
\scalebox{\tablescaler}{
\begin{tabular}{l cc @{\hskip 8pt} cccc @{\hskip 8pt} cc @{\hskip 8pt} c}
\toprule
& \multicolumn{2}{c}{Challenge metrics} & \multicolumn{4}{c}{Sub-metrics}
& \multicolumn{2}{c}{ADE@3\,s (m)} & \\
\cmidrule(lr){2-3} \cmidrule(lr){4-7} \cmidrule(lr){8-9}
& Scene & km\,/ & Coll.\ at-fault & Off-road & Corridor & Progress & NuRec & nuPlan & Latency \\
Method & score (\%) $\uparrow$ & infr.\ $\uparrow$ & (\%) $\downarrow$ & (\%) $\downarrow$
& (\%) $\downarrow$ & (\%) $\uparrow$ & (sim) $\downarrow$ & (real) $\downarrow$ & (ms) $\downarrow$\\
\midrule
LTFv6 Base~\cite{Nguyen2026CVPR} & 26.5{\scriptsize\,$\pm$\,0.1} & 0.23{\scriptsize\,$\pm$\,0.00} & \textbf{10.2{\scriptsize\,$\pm$\,0.4}} & 20.7{\scriptsize\,$\pm$\,0.0} & 27.6{\scriptsize\,$\pm$\,0.2} & 43.7{\scriptsize\,$\pm$\,0.0} & 4.34{\scriptsize\,$\pm$\,0.00} & 0.74{\scriptsize\,$\pm$\,0.00} & \multirow{4}{*}{23} \\
\quad + BC & 12.8{\scriptsize\,$\pm$\,2.1} & 0.24{\scriptsize\,$\pm$\,0.02} & 26.0{\scriptsize\,$\pm$\,1.1} & 21.3{\scriptsize\,$\pm$\,2.3} & 22.3{\scriptsize\,$\pm$\,1.3} & 50.9{\scriptsize\,$\pm$\,1.9} & \textbf{0.60{\scriptsize\,$\pm$\,0.00}} & \textbf{0.53{\scriptsize\,$\pm$\,0.01}} &  \\
\quad + RoaD-recovery~\cite{Garcia-Cobo2026CVPRFIND} & 32.9{\scriptsize\,$\pm$\,3.2} & 0.42{\scriptsize\,$\pm$\,0.05} & 17.1{\scriptsize\,$\pm$\,1.9} & 14.0{\scriptsize\,$\pm$\,1.0} & \textbf{10.6{\scriptsize\,$\pm$\,2.0}} & 55.0{\scriptsize\,$\pm$\,4.3} & 0.75{\scriptsize\,$\pm$\,0.01} & 0.72{\scriptsize\,$\pm$\,0.01} &  \\
\quad + OPTED (ours) & \textbf{41.8{\scriptsize\,$\pm$\,0.4}} & \textbf{0.43{\scriptsize\,$\pm$\,0.01}} & 23.4{\scriptsize\,$\pm$\,0.7} & \textbf{11.7{\scriptsize\,$\pm$\,1.0}} & 18.6{\scriptsize\,$\pm$\,0.5} & \textbf{73.4{\scriptsize\,$\pm$\,0.2}} & 1.67{\scriptsize\,$\pm$\,0.05} & 1.22{\scriptsize\,$\pm$\,0.06} &  \\
\addlinespace
VaVAM Base~\cite{Bartoccioni2025ARXIV} & 3.9{\scriptsize\,$\pm$\,0.2} & 0.08{\scriptsize\,$\pm$\,0.00} & 18.6{\scriptsize\,$\pm$\,0.2} & 34.1{\scriptsize\,$\pm$\,0.5} & 48.6{\scriptsize\,$\pm$\,1.0} & 31.2{\scriptsize\,$\pm$\,0.2} & 6.78{\scriptsize\,$\pm$\,0.01} & 2.78{\scriptsize\,$\pm$\,0.00} & \multirow{4}{*}{112} \\
\quad + BC & 25.2{\scriptsize\,$\pm$\,1.1} & 0.18{\scriptsize\,$\pm$\,0.02} & 12.8{\scriptsize\,$\pm$\,0.8} & 32.1{\scriptsize\,$\pm$\,2.0} & 32.1{\scriptsize\,$\pm$\,1.3} & 53.8{\scriptsize\,$\pm$\,1.6} & \textbf{4.43{\scriptsize\,$\pm$\,0.15}} & \textbf{1.93{\scriptsize\,$\pm$\,0.01}} &  \\
\quad + RoaD-recovery~\cite{Garcia-Cobo2026CVPRFIND} & 30.7{\scriptsize\,$\pm$\,4.2} & 0.22{\scriptsize\,$\pm$\,0.02} & \textbf{11.0{\scriptsize\,$\pm$\,1.1}} & 27.9{\scriptsize\,$\pm$\,1.1} & 28.6{\scriptsize\,$\pm$\,3.6} & 55.6{\scriptsize\,$\pm$\,2.2} & 4.92{\scriptsize\,$\pm$\,0.08} & 2.14{\scriptsize\,$\pm$\,0.11} &  \\
\quad + OPTED (ours) & \textbf{37.1{\scriptsize\,$\pm$\,2.1}} & \textbf{0.25{\scriptsize\,$\pm$\,0.03}} & 13.5{\scriptsize\,$\pm$\,0.7} & \textbf{24.8{\scriptsize\,$\pm$\,1.9}} & \textbf{28.4{\scriptsize\,$\pm$\,1.4}} & \textbf{60.7{\scriptsize\,$\pm$\,1.7}} & 6.20{\scriptsize\,$\pm$\,0.05} & 3.40{\scriptsize\,$\pm$\,0.18} &  \\
\addlinespace
Qwen-Drive 1.0~\cite{Zhou2026ARXIV} & 25.4{\scriptsize\,$\pm$\,0.5} & 0.21{\scriptsize\,$\pm$\,0.01} & 14.4{\scriptsize\,$\pm$\,0.7} & 18.3{\scriptsize\,$\pm$\,1.1} & 19.0{\scriptsize\,$\pm$\,0.7} & 39.1{\scriptsize\,$\pm$\,0.2} & 1.09{\scriptsize\,$\pm$\,0.00} & 0.31{\scriptsize\,$\pm$\,0.00} & 1181 \\
Alpamayo 1.5~\cite{Wang2025ARXIV} & 55.2{\scriptsize\,$\pm$\,1.0} & 0.92{\scriptsize\,$\pm$\,0.05} & 4.8{\scriptsize\,$\pm$\,0.6} & 12.0{\scriptsize\,$\pm$\,1.4} & 25.5{\scriptsize\,$\pm$\,0.9} & 76.8{\scriptsize\,$\pm$\,0.8} & 0.57{\scriptsize\,$\pm$\,0.00} & 3.31{\scriptsize\,$\pm$\,0.01} & 940 \\
\bottomrule
\end{tabular}
}
\vspace{-0.5cm}
\end{table*}

\subsection{Problem Formulation}
\label{sec:method:problem}

We consider navigation in recorded real-world driving scenes, where the policy must control the ego vehicle to complete a route while avoiding collisions and staying on the road.
At time $t$ the policy $\pi_\theta(\cdot \mid o_t)$ maps an observation $o_t=o(s_t)$ of scene state $s_t$, either vectorized (e.g., an HD map) or raw sensor data, to a motion decision. We consider two common output heads. Action policies emit a control action $a_t$ applied through the vehicle dynamics. Trajectory policies emit waypoints $\tau_t=(\mathbf{w}_1,\dots,\mathbf{w}_H)$, $\mathbf{w}_k=(x_k,y_k)$ which a tracking controller $C$ converts to actions, $a_t=C(s_t,\tau_t)$. Execution is closed-loop: each action advances the scene to $s_{t+1}$, where the policy is queried again. Policies pre-trained open-loop on human demonstrations can degrade in this regime: errors can compound into states not well covered by demonstrations~\cite{Ross2010AISTATS}. Our goal is to restore closed-loop performance while minimizing rendered interaction.

We distinguish two variants of closed-loop training: RL, which optimizes return under the policy’s own state distribution but typically requires orders of magnitude more interaction than supervised learning~\cite{Sun2017ICML,Cusumano-Towner2025ICML}; and on-policy supervision in the DAgger family~\cite{Ross2011AISTATS}, that iteratively executes the current policy, labels visited states with an expert, aggregates them into the training set, and retrains the policy. Unlike behavior cloning, the imitation loss for DAgger is minimized over the learner’s closed-loop state distribution:
\begin{equation}
    \min_\theta \;\; \mathbb{E}_{s \sim d_{\pi_\theta}}
    \Big[\, \ell\big(\pi_\theta(\cdot \mid o(s)),\; y^{*}(s)\big) \,\Big]
    \label{eq:opd}
\end{equation}
with target $y^{*}(s) = \Lambda\big(\pi^{*}, s\big)$, where $d_{\pi_\theta}$ is the finite-horizon state distribution induced by $\pi_\theta$, $\pi^{*}$ is the expert policy queried at the visited state, and $\Lambda$ converts its output to the student’s interface. With $\Lambda=\mathrm{id}$, Eq.~\eqref{eq:opd} reduces to the DAgger objective.

OPTED instantiates Eq.~\eqref{eq:opd} with a learned vectorized RL teacher as $\pi^{*}$ (Sec.~\ref{sec:method:teacher}), action-to-trajectory conversion as $\Lambda$ for waypoint-predicting students, and a loss $\ell$ matched to the student's output head: KL divergence between action distributions or trajectory regression against the converted plan (Sec.~\ref{sec:method:opd}). The student can be any policy pre-trained open loop on human data, with no architectural or output-head constraints beyond the interface of Eq.~\eqref{eq:opd}.

\subsection{Privileged Teacher Training}
\label{sec:method:teacher}

The expert of Eq.~\eqref{eq:opd} is a privileged policy trained with PPO~\cite{Schulman2017ARXIV} in PufferDrive~2.0~\cite{Cornelisse2025GH}, a vectorized render-free simulator. It observes the privileged vectorized state, namely ego kinematics, surrounding agents' poses and velocities, and the local road graph, and emits a continuous control action. We depart from the PufferDrive~2.0 system in two respects, the policy architecture and the reward, to obtain a teacher that drives smoothly and at human-like speed.
The released policy encodes each observation block with max-pooled MLPs feeding an LSTM.
Instead, we use an attention encoder in which the ego embedding cross-attends separately to agent and map-element tokens, with no recurrence. The released reward is sparse, a terminal goal bonus plus per-step collision and off-road penalties, so that an episode earns positive reward only if the goal is reached, which favors fast driving. We augment it with a dense route-progress reward gated on lane alignment, so partial progress along the route is rewarded without requiring reaching the goal:
\begin{equation}
      r_t = \mathbb{I}_{\mathrm{lane},t}\, \Delta \mathrm{RC}_t + w_{g}\,\mathbb{I}_{\mathrm{goal},t} - w_{c}\,\mathbb{I}_{\mathrm{coll},t} - w_{o}\,\mathbb{I}_{\mathrm{off},t}
    \label{eq:teacher-reward}
\end{equation}
Here the route is the logged human trajectory of the scenario, with total arc length $L$. At each step the ego position is projected onto the nearest point of the route, and $d_t$ denotes the arc length from the route's start to this projected point, with the projection admitted only while the ego is within a distance $D_{\max}$ of the route. Route completion is the running maximum $\mathrm{RC}_t = \max_{t' \leq t} d_{t'} / L \in [0, 1]$, so the increment $\Delta \mathrm{RC}_t = \mathrm{RC}_t - \mathrm{RC}_{t-1}$ is non-negative by construction and the progress term sums to at most one over an episode. The projection is spatial rather than temporal, rewarding the model for advancing along the route independently of its speed. The lane gate $\mathbb{I}_{\mathrm{lane},t}$ grants progress only while the ego is within a fixed distance of the nearest lane centerline and within a fixed angle of its tangent, and the goal term is a small recurring bonus for reaching the log waypoint a fixed time ahead. The per-step safety penalties do not terminate the episode, so the teacher can accumulate multiple penalties while driving in infraction states and learns to recover from them.
Our teacher is first trained on a large corpus of vectorized driving scenes and then fine-tuned, still render-free, on the target scene distribution on which the student is fine-tuned. After this fine-tuning, the teacher is frozen and used to label student states during distillation.

\subsection{On-Policy Distillation}
\label{sec:method:opd}

\paragraph{Distillation loop}
OPTED follows the iterative structure of DAgger, alternating data collection and policy updates. The student drives closed-loop in the target end-to-end simulator and the states it visits are labeled by the frozen teacher at collection time. The resulting state-label pairs enter a bounded FIFO buffer of recent rollouts, from which the current student is updated by minimizing Eq.~\eqref{eq:opd}. The updated student then collects new rollouts in the new training iteration. The term \emph{on-policy} signifies that supervision is anchored at states the student itself visits, with the target at each state being the teacher-generated future plan from it. Because updates draw only on recent rollouts, the training distribution tracks $d_{\pi_\theta}$ as the student improves.
Equation~\eqref{eq:opd} does not constrain which parameters are updated. The sets of trained parameters in each setting are stated in Sec.~\ref{sec:exp}.

\paragraph{Teacher labeling ($\Lambda$)}
Given the state $s_t$ visited by the student, $\Lambda$ places the teacher at $s_t$ and lets it drive for $K$ steps with its mean actions, over bicycle-model dynamics and log-replayed traffic. Since the teacher observes the vectorized state, this rollout costs no rendering. The positions it drives through, resampled to the student's prediction times in its planning frame at $t$, form the target plan $\tau^{*}_t = (\mathbf{w}^{*}_1, \dots, \mathbf{w}^{*}_H)$. The anchor state is the student's and the future is the teacher's.

\paragraph{Output-matched losses ($\ell$)}
The loss follows the student's head, so one teacher supervises students regardless of their output interface. For action heads ($\Lambda = \mathrm{id}$), used for the action-space students of Sec.~\ref{sec:exp:womd}, $\ell$ is the forward KL divergence from the teacher's to the student's action distribution:
\begin{equation}
  \ell_{\mathrm{KL}} \;=\; D_{\mathrm{KL}}\big(\pi^{*}(\cdot \mid s_t) \,\|\, \pi_\theta(\cdot \mid o_t)\big)
  \label{eq:kl-loss}
\end{equation}
For trajectory regressors, $\ell$ is an $\mathrm{L}_1$ regression of the predicted plan onto the converted plan $\tau^{*}_t$, the negative log-likelihood of a fixed-scale Laplace student and thus the distillation objective in its deterministic limit:
\begin{equation}
  \ell_{\mathrm{traj}} \;=\; \sum_{k=1}^{H} \alpha_k \, \lVert \mathbf{w}_k - \mathbf{w}^{*}_k \rVert_1
  \label{eq:traj-loss}
\end{equation}
The weights $\alpha_k$ emphasize the part of the prediction horizon that the tracking controller $C$ consumes, so that supervision concentrates on the waypoints that determine the executed action.

\section{Experiments}
\label{sec:exp}

\subsection{End-to-End Closed-Loop Fine-Tuning}
\label{sec:exp:end-to-end}

\begin{figure*}[t]
\centering
\includegraphics[width=\textwidth]{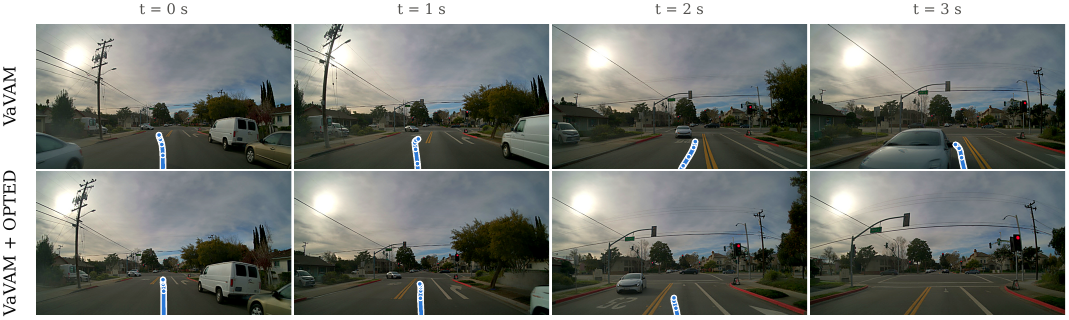}
\caption{\textbf{Qualitative results of closed-loop rollouts in AlpaSim.} \textbf{Top:} the base VaVAM fails to stay in its lane while approaching an intersection and collides with an oncoming vehicle. \textbf{Bottom:} VaVAM after OPTED reaches the intersection and correctly stops at the red traffic light.}
\label{fig:qualitative}
\vspace{-0.5cm}
\end{figure*}

Our experiments evaluate how fine-tuning with OPTED improves pre-trained end-to-end driving models.
We evaluate against a range of baselines, including fine-tuning strategies using other DAgger-like approaches, and applying RL directly to the student.
We find that OPTED outperforms other fine-tuning strategies, reaching performance on par with RL, but at a fraction of the cost. 
We also compare against large-scale vision-language-action (VLA) reference policies and investigate how the student's output representation, as well as the teacher's model architecture affect the student's policy performance (Sec.~\ref{sec:exp:womd}).

All end-to-end experiments are conducted in AlpaSim~\cite{alpasim_2025}.
Two pre-trained driving models are fine-tuned and evaluated on a held-out split of the public PAI-AV NuRec dataset~\cite{NVIDIA2025HF}, as well as on the AlpaSim end-to-end challenge leaderboard~\cite{NVIDIA2026HF}. 
Some ablation studies, in particular those requiring RL training, are conducted in PufferDrive~\cite{Cornelisse2025GH} on WOMD~\cite{Ettinger2021ICCV}. 
All results are reported as mean and standard deviation over three training seeds, evaluated at the final checkpoint.
As base and reference policies only provide a single checkpoint, we instead report the statistics over three independent evaluation runs.

\paragraph{Simulator and data}
AlpaSim renders NuRec 3D Gaussian splatting reconstructions of real-world driving scenes, which include high-speed highway driving as well as dense urban and rural settings. Each scene consists of a 20\,s clip with non-reactive log-replayed traffic. 
Simulation operates at 10\,Hz, including sensor input rendering and policy inference.  
The navigation commands are provided as one-hot encoded `left', `straight' or `right' signals which are derived from the logged trajectory 40\,m ahead of the ego.

We use 1669 scenes for training and 441 scenes from the public challenge split for evaluation. For evaluation in PufferDrive, we were only able to use a subset of 420 scenes due to data incompatibility between the Physical AI dataset and the PufferDrive format. 

\paragraph{Metrics}
We report the two main metrics of AlpaSim, computed per scene $i\in[1\dots N]$.
Let $p_i \in [0,1]$ be the projected route progress along the logged trajectory and $d_i$ the distance in kilometers driven up to that point. A scene \emph{fails} ($f_i = 1$) on an at-fault collision, an off-road event, or a lateral exit from the 4\,m corridor around the logged trajectory\footnote{This condition prevents evaluation in states with poor rendering quality.}. 
The \emph{scene score} saturates for any scene with progress of at least 80\% and is zero for failed scenes. \emph{Km/infr.} is the total distance driven per infraction, where an infraction ($n_i = 1$) is an at-fault collision or off-road event:
\begin{equation}
\begin{gathered}
  \mathrm{Scene\ score} = \frac{1}{N}\sum_{i} (1-f_i)\,\min\!\Big(\frac{p_i}{0.8},\,1\Big) \\
  \mathrm{km/infr.} = \frac{\sum_{i} d_i}{\sum_{i} n_i}
\end{gathered}
\label{eq:metrics}
\end{equation}
Note that metrics, including progress, are only aggregated along each trajectory until the first collision, off-road or off-corridor event occurs.

We also report the sub-metrics of at-fault collision rate, off-road rate, corridor exits and route progress. 
At-fault collisions are front and lateral impacts only, due to the non-reactive nature of log-replay traffic simulation. 
As a measure of open-loop agreement with human driving, we report the average displacement error (ADE) at 3\,s between the predicted plan and the logged trajectory on the NuRec evaluation scenes and on nuPlan~\cite{Caesar2021CVPRWORK}, the pre-training dataset of both students. Finally, we report inference latency per planning step on one NVIDIA H100 GPU.

\paragraph{Teacher}
The teacher is the policy of Sec.~\ref{sec:method:teacher}, pre-trained with PPO~\cite{Schulman2017ARXIV} on WOMD (Sec.~\ref{sec:exp:womd}) and then adapted to the PAI-AV dataset~\cite{NVIDIA2025HFPAI} map format by converting its scenes to the PufferDrive format and continuing PPO for 500M render-free environment interactions. Reward and PPO configuration are unchanged, except that the route goal is resampled every 4\,s to cover the 20\,s scenes (on WOMD, one route goal at the end of each 9\,s trajectory is provided). 
The same teacher is used across experiments in Table~\ref{tab:e2e_main_results}. 

\paragraph{Student models}
LTFv6~\cite{Nguyen2026CVPR} is pre-trained with behavior cloning on nuPlan and CARLA~\cite{Dosovitskiy2017CORL} data from a privileged planner. It also consumes ego speed and acceleration in addition to camera input. VaVAM~\cite{Bartoccioni2025ARXIV} is a video model pre-trained on OpenDV~\cite{Yang2024CVPR} with a flow-matching waypoint head trained on nuPlan and nuScenes~\cite{Caesar2020CVPR} from a single front camera. We integrate the flow-matching head from a fixed noise sample, so both students are deterministic. We train only the planning decoder and freeze the backbone, to avoid overfitting to rendering artifacts and allow caching backbone latents in the FIFO buffer (Sec.~\ref{sec:method:opd}). Both models re-plan at 10\,Hz over their native horizon, 8 and 6 waypoints respectively at 0.5\,s spacing, tracked by the default AlpaSim MPC controller.

\paragraph{Fine-tuning variants}
All variants train the planning decoder for two epochs on the training scenes. \emph{Base} denotes the released checkpoint. \emph{BC} denotes open-loop fine-tuning on logged trajectories from the training scenes re-rendered in AlpaSim. This setting separates domain adaptation from closed-loop gains. \emph{RoaD-recovery} denotes closed-loop supervised fine-tuning with a log-anchored target, our re-implementation of the recovery mode of RoaD~\cite{Garcia-Cobo2026CVPRFIND} within the closed-loop SFT framework of CAT-K~\cite{Zhang2025CVPR}. These methods execute the candidate closest to the log among several sampled per step, which does not transfer to a policy that emits a single deterministic plan.
Accordingly, we cannot apply the full RoaD algorithm, but only utilize its ``recovery mode'': whenever the student's plan departs from the logged continuation by more than 3\,m, the executed plan interpolates from the prediction onto the log over the following 2\,s, and the recovery trajectory is the supervision target at every step.
RoaD-recovery therefore differs from OPTED in what is executed and supervised: 
RoaD-recovery's rollouts are biased towards the logged trajectories while OPTED rollouts are unconstrained. Additionally, RoaD-recovery's training targets are constructed by manual heuristics and often exhibit sub-optimal or unsafe behavior.
Both closed-loop variants minimize the weighted L1 loss of Sec.~\ref{sec:method:opd} on a FIFO buffer of the 512 most recent rollouts, 32 of which are collected by the current student at each iteration.

\begin{table}[t]
\centering
\caption{Submissions to the official AlpaSim challenge leaderboard.}
\label{tab:challenge_leaderboard}
\scalebox{\tablescaler}{
\begin{tabular}{lrr}
\toprule
Policy & Scene Score (\%) $\uparrow$ & km\,/\,infr.\ $\uparrow$ \\
\midrule
LTFv6~\cite{Nguyen2026CVPR} & 25.1 & 0.14 \\
\quad + OPTED (ours) & 32.2 & 0.21 \\
VaVAM~\cite{Bartoccioni2025ARXIV} & 6.7 & 0.07 \\
\quad + OPTED (ours) & \textbf{35.8} & \textbf{0.22} \\
\bottomrule
\end{tabular}
}
\vspace{-0.5cm}
\end{table}

\paragraph{Main results}
Table~\ref{tab:e2e_main_results} reports the main AlpaSim results on the public NuRec evaluation split. At equal training budget, OPTED is the best fine-tuning variant on both examined architectures and both challenge metrics, raising the scene score of LTFv6 from 26.5 to 41.8 and of VaVAM from 3.9 to 37.1, with km/infr. improving alongside. The two baselines separate the sources of these gains. BC trains on the same scenes with rendered inputs, but only at logged states.
Its improvements, in particular for VaVAM, thus highlight the contribution of adapting the student to rendered inputs and the scene distribution used for evaluation. 
Importantly, the additional performance improvements of RoaD-recovery and OPTED, compared to BC, are due to training the student on closed-loop rollouts.
Lastly, the improved performance of OPTED over RoaD-recovery is explained by its unbiased rollouts and better supervision signal.
Note that while OPTED appears to increase at-fault collisions for LTFv6, the reported value is unnormalized by the distance traveled. Accounting for it, the distance between collisions is comparable to the baseline. At the same time, off-road events are nearly halved, even before taking distance normalization into account, explaining the greatly improved scene score and km/infr.
Fig.~\ref{fig:qualitative} shows a qualitative example: VaVAM leaves its lane before an intersection and collides with oncoming traffic, while the same model after OPTED reaches the intersection and correctly stops at the red light.

\paragraph{Comparison with large VLA models}
As a reference, we include two large-scale VLA models in Table~\ref{tab:e2e_main_results}: Qwen-Drive~1.0~\cite{Zhou2026ARXIV} and Alpamayo~1.5~\cite{Wang2025ARXIV}. Somewhat surprisingly, OPTED post-trained policies outperform Qwen-Drive, though they still lag behind Alpamayo. Note that both of these VLAs are significantly larger than our student policies (5.1--10B vs. 62--356M parameters); they were pre-trained on significantly more data; and their inference times (940--1181\,ms) make them unsuitable for real-time execution.

\paragraph{Open-loop agreement with human driving}
The ADE columns of Table~\ref{tab:e2e_main_results} measure open-loop agreement with the human trajectory in the target domain (NuRec) and in the pre-training domain of the students (nuPlan). Open-loop and closed-loop performance are decoupled: BC fine-tuning minimizes both ADEs for both students and has the lowest scene score of the fine-tuning methods. Qwen-Drive has the lowest nuPlan ADE in the table (it was trained on nuPlan data), yet a lower scene score than both OPTED students. As expected, OPTED does result in higher ADE than BC and RoaD-recovery, given its RL teacher targets, but the drift is bounded.
Fine-tuned students stay reasonably close to human driving while gaining closed-loop competence, in agreement with the smoothness result of Sec.~\ref{sec:exp:womd}.

\paragraph{AlpaSim challenge}
Table~\ref{tab:challenge_leaderboard} reports our submissions to the AlpaSim E2E challenge, evaluated on a private scene set. OPTED improves LTFv6 from 25.1 to 32.2 and VaVAM from 6.7 to 35.8 in scene score, so both improvements transfer to the public benchmark.

\paragraph{Teacher evaluation}
We evaluate the labels generated by the teacher in AlpaSim, in the same setting in which end-to-end policies are evaluated. It achieves a scene score of 64.8\% and 0.78 km/infr, compared with scores of 41.8\% and 37.1\% for its OPTED students. 
This gap has three possible sources: the non-matching output representation of the student and teacher, whose effect is isolated in Sec.~\ref{sec:exp:womd}; the non-matching input modalities (rendered images vs. privileged state); and limited transfer capacity of our distillation setup that, among others, only optimizes the decoder for two epochs. On the other hand, the gap to a perfect score indicates more improvement opportunity with a stronger teacher. Further, when the teacher is executed with its native actions, instead of $\Lambda$ conversion and downstream controller, it achieves 1.75 km/infr., suggesting better attainable gains for action-output student policies.

\begin{figure}[t]
\centering
\includegraphics[width=\columnwidth]{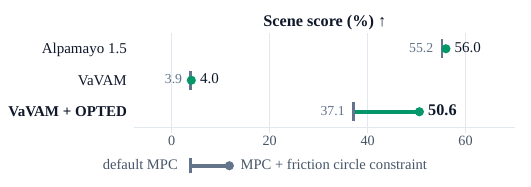}
\caption{\textbf{The effect of controller choice.} We find that adding a friction circle constraint to the default MPC controller can improve driving significantly. It prevents aggressive swerves when the target velocity is infeasible, a typical failure case for VaVAM which does not take the current speed as input.}
\label{fig:controller_ablation}
\vspace{-0.5cm}
\end{figure}

\paragraph{Controller analysis}
Beyond covariate shift, a waypoint plan is not guaranteed to be dynamically consistent with the current vehicle state, and when it is not, the outcome is determined by the tracking controller. VaVAM does not observe the ego speed and is pre-trained on slower scenes: fewer than 1\% of nuScenes and nuPlan scenes exceed 50\,km/h, against 22\% of NuRec scenes. At highway speed it predicts plans slower than the current state, and the default controller, which bounds longitudinal but not lateral acceleration, converts the implied braking into lateral swerving and off-road events. Constraining both accelerations to a friction circle~\cite{Rajamani2012} at evaluation time lowers the off-road rate of VaVAM before and after OPTED and raises the scene score of the latter from 37.1 to 50.6 (Fig.~\ref{fig:controller_ablation}), whereas Alpamayo, which observes the current speed, is practically unaffected, so the failure originates in the plan rather than in the controller.

\paragraph{Computational cost}
On an 8-GPU NVIDIA H100 node, one OPTED iteration with VaVAM spends 524\,s collecting 32 scenes of 20\,s in AlpaSim, 13.5\,s on teacher labeling and 6\,s on gradient updates over the 512 buffered scenes. Rendering and student inference dominate runtime, two epochs take 17\,h. The teacher is trained once on one NVIDIA RTX 4090 GPU in about 24\,h, plus 6\,h of PAI-AV NuRec adaptation, and is reused for every student.

\begin{figure}[t]
\centering
\includegraphics[width=\columnwidth]{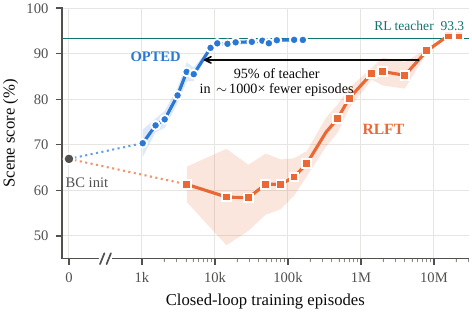}
\caption{\textbf{Sample efficiency of OPTED against RL fine-tuning (RLFT)}, both initialized from the same BC-pre-trained action student. Scene score on the WOMD test split against closed-loop episodes consumed during training. The arrow spans the episodes needed to reach 95\% of the teacher's scene score: 6.8k for OPTED vs.\ 6.2M for RLFT.}
\label{fig:success_vs_episodes}
\vspace{-0.5cm}
\end{figure}


\subsection{Vectorized Student Experiments on WOMD}
\label{sec:exp:womd}

We perform additional experiments where both teacher and student receive vectorized inputs, to answer questions that the end-to-end setting does not permit: how sample efficient is OPTED against RL fine-tuning of the same student; what is the effect of the student's output representation (waypoints vs. actions); and what is the effect of the teacher's design on the distilled student.

\paragraph{Setup and metrics}
We use PufferDrive on WOMD~\cite{Ettinger2021ICCV} as the target environment. Each scenario is a 9\,s segment at 10\,Hz with one controlled ego agent, log-replayed traffic and bicycle-model dynamics. All models observe the vectorized ego state, the nearest 200 road-graph elements, 63 agents with states and bounding boxes, and the route goal, so there is no observability gap between teacher and student. The teacher of Sec.~\ref{sec:method:teacher} is trained from scratch with PPO for 2B environment interactions, 22M closed-loop episodes, on 80k training scenes with the reward of Eq.~\eqref{eq:teacher-reward}. Two students are pre-trained with BC on the human demonstrations of the same split: an action student with Gaussian acceleration and steering outputs, as the teacher, and a waypoint student tracked by a re-implementation of the AlpaSim MPC controller. Distillation updates all parameters with Eq.~\eqref{eq:kl-loss} for the former and Eq.~\eqref{eq:traj-loss} for the latter. RL fine-tuning (RLFT) continues PPO from the BC action student with the teacher's configuration unchanged. Results are on the 10000 scenes of the WOMD test split, with scene score and km/infr.\ as in Eq.~\eqref{eq:metrics} but without the corridor condition. To test whether fine-tuning alters the driving comfort inherited from human data, we report the Wasserstein-1 distance in m/s$^3$ between the distributions of absolute longitudinal and lateral ego jerk, pooled over all steps and scenes, and those of the logged human trajectories ($\mathcal{W}_{\mathrm{lon}}$, $\mathcal{W}_{\mathrm{lat}}$).

\paragraph{Closed-loop fine-tuning sample efficiency}
Fig.~\ref{fig:success_vs_episodes} plots the scene score of the action student against closed-loop training episode count for OPTED and RLFT from the same BC initialization. OPTED reaches 95\% of the teacher's scene score after 6.8k episodes, roughly three orders of magnitude fewer than RLFT at 6.2M. Both converge to the teacher's level (Table~\ref{tab:womd}), OPTED at 93.0 after 160k episodes, and RLFT at 93.9 after 22M, where it also exceeds the teacher in km/infr.\ by optimizing the reward directly. RLFT requires millions of episodes even with vectorized inputs, which would have to be rendered by a sensor-based student. The advantage of OPTED is that it requires only on-policy imitation in the rendered simulator and moves the exploration into the render-free teacher.

\begin{table}[t]
\centering
\caption{Closed-loop evaluation of vectorized policies on the WOMD test split. $\mathcal{W}_{\mathrm{lon}}$/$\mathcal{W}_{\mathrm{lat}}$: Wasserstein-1 distance (m/s$^3$) between the policy-based and the human distributions of absolute longitudinal/lateral jerk. Bold marks the best value per student architecture.}
\label{tab:womd}
\scriptsize
\setlength{\tabcolsep}{2pt}
\resizebox{\columnwidth}{!}{%
\begin{tabular}{lcccc}
\toprule
Method & \shortstack{Scene\\Score (\%) $\uparrow$} & \shortstack{km/\\infr.\ $\uparrow$} & $\mathcal{W}_{\mathrm{lon}}\,\downarrow$ & $\mathcal{W}_{\mathrm{lat}}\,\downarrow$ \\
\midrule
\multicolumn{5}{l}{\textit{RL teachers}} \\
\quad LSTM, sparse~\cite{Cornelisse2025GH} & 97.4{\scriptsize$\pm$0.4} & 2.66{\scriptsize$\pm$0.37} & 18.46{\scriptsize$\pm$0.07} & 17.27{\scriptsize$\pm$2.14} \\
\quad LSTM, dense & 97.7{\scriptsize$\pm$0.2} & 3.74{\scriptsize$\pm$0.14} & 13.31{\scriptsize$\pm$0.40} & 13.53{\scriptsize$\pm$1.41} \\
\quad Ours (attn., dense) & 93.3{\scriptsize$\pm$0.7} & 3.17{\scriptsize$\pm$0.12} & 1.10{\scriptsize$\pm$0.28} & 3.14{\scriptsize$\pm$0.16} \\
\midrule
Action student & 68.2{\scriptsize$\pm$3.1} & 0.25{\scriptsize$\pm$0.04} & 3.77{\scriptsize$\pm$0.45} & \textbf{0.91}{\scriptsize$\pm$0.03} \\
\quad + RLFT & 93.9{\scriptsize$\pm$0.2} & \textbf{3.47}{\scriptsize$\pm$0.27} & 1.89{\scriptsize$\pm$0.10} & 3.26{\scriptsize$\pm$0.05} \\
\quad + OPTED (LSTM, sparse) & 96.6{\scriptsize$\pm$0.2} & 1.75{\scriptsize$\pm$0.10} & 19.19{\scriptsize$\pm$0.09} & 13.14{\scriptsize$\pm$0.63} \\
\quad + OPTED (LSTM, dense) & \textbf{96.8}{\scriptsize$\pm$0.3} & 2.08{\scriptsize$\pm$0.16} & 14.63{\scriptsize$\pm$0.43} & 7.90{\scriptsize$\pm$0.35} \\
\quad + OPTED (ours) & 93.0{\scriptsize$\pm$0.1} & 2.86{\scriptsize$\pm$0.09} & \textbf{1.00}{\scriptsize$\pm$0.01} & 2.04{\scriptsize$\pm$0.02} \\
\addlinespace
Waypoint student & 73.8{\scriptsize$\pm$1.8} & 0.43{\scriptsize$\pm$0.02} & \textbf{4.29}{\scriptsize$\pm$0.35} & \textbf{0.10}{\scriptsize$\pm$0.01} \\
\quad + OPTED (ours) & \textbf{87.8}{\scriptsize$\pm$0.3} & \textbf{1.33}{\scriptsize$\pm$0.04} & 5.02{\scriptsize$\pm$0.01} & 0.70{\scriptsize$\pm$0.01} \\
\bottomrule
\end{tabular}}
\vspace{-0.5cm}
\end{table}

\paragraph{Output representation and driving comfort}
From the same teacher, OPTED improves both students in scene score and km/infr.\ (Table~\ref{tab:womd}), the action student by more than the waypoint student. This validates the waypoint path deployed in Sec.~\ref{sec:exp:end-to-end} while showing that the output mismatch limits transferred competence, although the two students are not directly comparable since the waypoint student drives through a tracking controller. The jerk distances are compared within each axis and read against the BC prior: the Wasserstein distance is symmetric, so the BC students, which drive more smoothly than the human drivers, are penalized for that deviation as a less smooth policy would be. At similar closed-loop competence, RLFT moves the action student onto the teacher's lateral jerk distribution and part of the way toward it longitudinally, whereas OPTED retains part of the BC lateral prior and brings the longitudinal distribution closer to the human one than the teacher itself.

\paragraph{Teacher design}
By observing the performance of the distilled student rather than the teacher's own return, we departed from the PufferDrive~2.0 recipe~\cite{Cornelisse2025GH} with the dense progress reward of Eq.~\eqref{eq:teacher-reward} and the non-recurrent attention architecture of Sec.~\ref{sec:method:teacher}. Under the sparse goal reward, which pays only for reaching the goal within the episode, the recurrent teacher's jerk distributions depart from the human ones by up to an order of magnitude more than ours (Table~\ref{tab:womd}), and the dense reward raises its km/infr.\ and reduces the lateral jerk distance of its student by 40\%. Both LSTM teachers exceed ours in scene score, which saturates with route progress and is indifferent to comfort, yet their students retain only 66\% and 56\% of their teacher's km/infr.\ against 90\% for ours, with jerk distances several times larger. The teacher that scores higher by its own metrics therefore produces the weaker student. We attribute this to two properties of the teacher as a labeler. First, distillation transfers driving style along with competence. Second, the teacher should be Markov: OPTED queries it at states produced by the student, so the label must be determined by the queried state alone. A recurrent teacher instead requires a hidden state the query does not provide, and its history-dependent decisions are not representable by a student acting on the current observation. The teacher should therefore be designed for what the student can imitate.

\section{Conclusion}
We presented OPTED, on-policy distillation of a privileged RL teacher into pre-trained end-to-end driving policies, in which reward-driven exploration is kept in a vectorized simulator and rendered interaction is spent only on supervision. On two policy architectures, closed-loop fine-tuning in AlpaSim improves driving scores significantly, more than alternatives, and the gains carry over to the private AlpaSim leaderboard. 
On WOMD, distillation reaches the teacher's performance level with ca.\ three orders of magnitude fewer episodes than RL fine-tuning, and resembles human driving more closely. Future work may improve our results by using a stronger privileged teacher, such as self-play policies at the scale of GigaFlow~\cite{Cusumano-Towner2025ICML}; or a stronger student policy, such as Alpamayo~\cite{Wang2025ARXIV}. We observed better teacher-student transfer with both policies using the same output representation, which motivates more research on end-to-end policies that predict actions rather than waypoints. Finally, the teacher observes the privileged state while the student perceives the scene through cameras. This observability gap could be addressed, e.g., as in LEAD~\cite{Nguyen2026CVPR}.

{
	\bibliographystyle{IEEEtran}
	\bibliography{bstctl,bibliography_short,bibliography_custom}
}
 
\end{document}